\documentclass{article}
\usepackage[xspace]{ellipsis}

\usepackage{iccc}
\usepackage[longnamesfirst]{natbib}
\usepackage[letterpaper,left=0.75in,width=7in,bottom=1.25in,top=0.75in]{geometry}
\usepackage{times}
\usepackage{helvet}
\usepackage{courier}
\title{Interdisciplinary Methods in Computational Creativity:\\ How Human Variables Shape Human-Inspired AI Research}
\author{Nadia M. Ady$^{*1,2}$ and Faun Rice$^{*3}$\\
$^1$Department of Computing Science, University of Alberta\\
$^2$Alberta Machine Intelligence Institute (Amii)\\
$^3$Independent Researcher\\
nmady@ualberta.ca, faun.rice@gmail.com
}
\begin{document} 

\maketitle

\def\thefootnote{*}\footnotetext{Both authors contributed equally to this work.}

\begin{abstract}
\begin{quote}
The word \textit{creativity} originally described a concept from human psychology, but in the realm of computational creativity (CC), it has become much more. The question of what creativity means when it is part of a computational system might be considered core to CC. Pinning down the meaning of creativity, and concepts like it, becomes salient when researchers port concepts from human psychology to computation, a widespread practice extending beyond CC into artificial intelligence (AI). Yet, the human processes shaping human-inspired computational systems have been little investigated. In this paper, we question \textit{which} human literatures (social sciences, psychology, neuroscience) enter AI scholarship and \textit{how} they are translated at the port of entry. This study is based on 22 in-depth, semi-structured interviews, primarily with human-inspired AI researchers, half of whom focus on creativity as a major research area. This paper focuses on findings most relevant to CC. We suggest that \textit{which} human literature enters AI bears greater scrutiny because ideas may become disconnected from context in their home discipline. Accordingly, we recommend that CC researchers document the decisions and context of their practices, particularly those practices formalizing human concepts for machines. Publishing reflexive commentary on human elements in CC and AI would provide a useful record and permit greater dialogue with other disciplines.
\end{quote}
\end{abstract}

\section{Introduction}
Computational creativity (CC) is informed by many human literatures, including psychology, sociology, cognitive science, and philosophy (\citealp[p.~11]{ackerman2017teaching}; \citealp{mcgregor2014computational}). There is a long history of reflection on the relationship between CC's parent, AI, with other disciplines \citep{newell1970remarks} which continues today (\citealp{lieto2016human}; \citealp{macpherson2021natural}; \citealp{cassenti2022editors}). Social sciences also offer relevant commentary: Science, Technology, and Society (STS) is concerned with how scientific methods produce knowledge and shape the world, calling attention to the human processes inherent in scientific work using a broad methodological toolkit (\citealp{jasanoff2013fields}; \citealp{lippert2021data}; \citealp{law2004after}; \citealp{suchman1993artificial}). In alignment with these conversations, our project explores the human processes involved when researchers draw inspiration from concepts from human psychology for computational systems. In this paper, we present early findings centred on CC. This work responds to calls to articulate the “methodological and conceptual barriers \ldots [which] confront attempts to work across disciplinary boundaries” (\citealp{macleod2018makes}, p.~697).

Our dataset is 22 in-depth, semi-structured interviews with CC and AI researchers working closely with concepts from human psychology (see Methodology). For 11 interviewees, the concept of creativity is a key thread in their research: the other 11 engaged with concepts such as curiosity, forgetting, or mental time travel. We use “human-inspired” as shorthand for this heterogeneous group throughout the paper, and transcripts from non-CC participants refine our understanding of each finding, though our focus here is on CC. We build on existing scholarship by suggesting that human and social factors impact \textit{which} human literature enters AI and \textit{how} it is translated for computation at its port of entry. Further, we suggest that human and social processes in CC are productive areas of inquiry, and that qualitative methods offer fruitful ways of exploring these topics, in agreement with scholars like \citet{perezyperez2020towards}. In demonstration, we outline two phenomena related to the challenges of interdisciplinary work, followed by an example of intellectual influence on human-inspired AI that emerged from qualitative interviews. 

\section{Methodology}

This study has used a grounded theory approach to conception, data collection, and analysis. Aligned with grounded theory methodologies, we began with a broad interest rather than a hypothesis (\citealp{qureshi2020paradigm}); prioritized inductive findings from primary qualitative research (\citealp{glaser1967discovery}); and participated collaboratively in transcription, line-by-line coding, memoing, focused coding, and forming early-stage conceptual categories (\citealp{wiener2007making}, p.~301; \citealp{charmaz2014constructing}). 

We began with purposive sampling of human-inspired CC and AI researchers. We used interviewees’ publications to assess their relevance to study aims, and proceeded via snowball sampling. In one-hour long semi-structured interviews, we asked participants how they defined the human concepts they worked with, what types of literature and personal experience had shaped their definitions, and what challenges or successes they encountered in translating their concepts for machines. 

Of our 22 participants, six used she/her pronouns, and all worked in North America or Europe; improved gender and regional diversity are goals of this study as we continue data collection. Participants included five employees of private sector AI firms, three PhD students, one postdoctoral researcher, four pre-tenure professors, and nine post-tenure professors. Academic interviewees were primarily in computational- or psychology-related departments. We use the convention P\# to anonymize participants. 

Grounded theory methodology espouses simultaneous data collection and analysis, suggesting that questions raised by early rounds of analysis should be pursued further in subsequent data collection as a form of theoretical sampling, alongside further review of relevant literature (\citealp{charmaz2014constructing}). Early dissemination through this paper allows us to incorporate diverse feedback into the study (\citealp{green2007grounded}, p.~489). Accordingly, we will further develop this project by taking cues from readers in the CC community, including additional data collection and deeper exploration of themes introduced in this paper. This project was approved by University of Alberta Research Ethics Board 1 (ID: Pro00109111). 

\section{How do Ideas from Human Literatures Enter CC and AI?}

The difficulty of reading broadly for interdisciplinary research was a core theme in interviews. Several interviewees felt that “understanding what is going on in all [the] different, relevant fields” was “one of [their] biggest challenges as researchers,” (P22). They often relied on “serendipity,” (P22) or popular culture: “I’m probably more likely to learn from a New York Times article profiling the day in the life of an artist than I am to actually read art history books” (P15). Others began conversations at conferences or across departments, or employed strategies like citation chaining to make headway. The precise language (jargon) required for high levels of rigor is known to sometimes prevent access by readers external to a discipline, meaning that this challenge may come with the territory of interdisciplinary work (\citealp{callaos2013interdisciplinary}, pp.~23-24; \citealp{daniel2022challenges} pp.~8-9; \citealp{macleod2018makes}, p.~707). 

As a result of this challenge, keeping up with ongoing debate and discussion in the human literatures may become deprioritized once an idea has gained traction in CC or AI. In some cases, knowledge about ideas' origins may be lost. P6 offered the example of catastrophic forgetting in machine learning, connecting it to the psychological term “retroactive inhibition,” and contending that the usage in machine learnin was initially congruent with the usage in psychology but became increasingly misaligned. Authors became less aware of the origins of the ideas they were citing over time. P6 sees a “strong disconnect” as a consequence: “the concept of forgetting as it appears in the psychological literature is much more broad and diverse than forgetting within artificial neural networks” (P6). 

The first time a concept is ported into AI from human literatures, it may be from a seminal scholar doing important translation work for their field. Margaret \citeauthor{boden2004creative}’s model of creativity as novelty, value, and surprise (\citeyear{boden2004creative}, p.~1), for example, was described by many as a “huge service” which “put forward ideas \ldots that had been around long before her work in the 1970s” but introduced them to cognitive and computer science (P12). Nevertheless, the fruits of translation work may still lose contact with ongoing discussions in other fields, and even ossify in CC or AI: for example, one interviewee suggested that citing Margaret \citeauthor{boden2004creative} had simply become part of the brand of CC (P21), and another recalled defaulting to citing her work when rebuked by “senior academics" for not "being specific in [their] definition” (P3). 

Similarly, scholars in other disciplines may pave the way for cross-disciplinary translation by synthesizing a topic in a popular non-fiction book or textbook. Such works can acquire a kind of virality in AI and its subfields. For example, multiple interviewees mentioned the works of \citeauthor{csikszentmihalyi1990flow} (e.g. \textit{Flow}, \citeyear{csikszentmihalyi1990flow}) and \citeauthor{tomasello2019becoming} (\textit{Becoming Human}, \citeyear{tomasello2019becoming}). P11 described \textit{Becoming Human }(2019) as “making the rounds among AI–psychology academics. It’s basically about what differentiates children from primates \ldots spanning development, psychology, primatology and others where it really hones in on what capabilities humans have \ldots And I just think that there are people that ask those questions and [\citeauthor{tomasello2019becoming}] presents them very clearly.” Textbooks and popular non-fiction offer essential knowledge translation, but come with constraints (e.g., editorial standards sometimes discourage authors from citing in-text; authors are offering a broad synthesis of ideas) that do not provide a full understanding of ongoing conversations in the authors’ home disciplines (\citealp{callaos2013interdisciplinary}, pp.~23-24).

We do not intend to undervalue influential cross-disciplinary contributions: interviewees saw such work as crucial and valuable, but simply warned of “tunnel vision” (P9) or “misalignment” (P6). When a single scholar becomes ‘the person’ to cite, it may permit a disjuncture between AI and ongoing debates in psychology or other disciplines. Other scholars have raised concerns about “herd mentality” (\citealp{rekdal2014academic}, p.~570) or maintenance of the status quo (\citealp{dworkin2020inciting}, p.~890) in contemporary citation practices. At the same time, citations help us position thinkers “as the authors, authorities and originators” and remember and acknowledge the genealogy of a field (\citealp{liu2021use}, p.~215) and our debts “to those who came before” (\citealp{ahmed2017living}, p.~15). Citing \citeauthor{boden2004creative} in particular, might be considered a feminist practice given the disproportionate number of men in AI, reminding the reader that CC is indebted to a woman scholar. 

Ironically, feeling that one has to adequately acknowledge an idea's disciplinary context may deter scholars from taking inventive paths to new knowledge–for example, one graduate student interviewee commented, “that's the difficulty with interdisciplinary work. You're never going to be just one or the other. You can't be the best at either field by devoting yourself to both, right?” (P1). \textbf{Yet, we would suggest that researchers should not be afraid to read broadly despite knowing that some disciplinary context will be lost. While key pieces of knowledge translation can generate valuable new lines of thinking, it is also important to recognize that no one source can explain a whole field. Instead, researchers can think through the effects of which sources from human literatures are influencing their work. }Reflecting on the influence of different ideas and disciplines might also allow CC researchers to identify gaps and opportunities for future research: for example, it is possible that work in English is more likely to make its way into CC. Similarly, ideas about creativity from disciplines beyond psychology and neuroscience (e.g., sociology, anthropology, philosophy) may fall outside of typical reading lists for historical and institutional reasons. 

\section{‘An Interpretation Job’: Articulating Hidden Methodological Decisions}

Interviewees described a second challenge related to translating ideas from human literatures to CC or AI. Many found concepts in “the human literature [to be] not well defined\ldots more of a nice metaphor” (P8), “open ended and ambiguous” (P22), or “extremely general” (P14). For example, interviewees described digging through work by Jean Piaget, Lev Vygotsky, or Daniel Berlyne seeking clarity on theoretical terminology. Others observed that many concepts might not yet have direct parallels in AI: “Appraisal theory of motivation, for instance, that’s very much connecting motivation with emotion or affect. And what does that even correspond to in AI? Now that’s quite a challenge. Is it even worth going down that route if we don’t have any corresponding elements for that in the computational domain?” (P22).  

As a result, some researchers attribute their choice of definition to ease of evaluation or formalization. Such decisions include discarding some definitions, like steering away from “\textit{H-creativity} [historical creativity], which is only theoretical, because how do you even measure [the idea] that no one in human history has ever had this thought before” (P3). Alternatively, P16 described looking through competing definitions in psychology and landing on one “easily translatable into reinforcement learning.” Finally, ease of evaluation might present an opportunity to contribute: P22 expressed enthusiasm for “modeling very minimal creativity because [they] think it is more amenable to measurement.”

While interviewees described some ideas from human literatures as more amenable to evaluation or formalization than others, processes of translation across that spectrum involved individual choices. We asked how interviewees “translated” definitions for computational systems, and one countered, “it’s not so much of a translation problem as that the first definition is blurry. It’s more of an interpretation type of job” (P16). P14 described a similar process:  “to do this translation of these psychological ideas, especially when they are ambiguous like most of them are, this process is what really makes the difference. It's not simple to do\ldots And there are a lot of things [in this process] that are so important for the development of knowledge and insight.” Articulating choices made during translation of psychological ideas for computer science not only helps future “translators,” it also tracks the changes in meaning that concepts may undergo during this process. 

Accordingly, \textbf{there are two stages where researchers might record decisions they make about using ideas from human literatures: first, at the point of selecting or discarding particular definitions, and second, during interpretation. }The use of reflexive (self-aware, positional) description of one’s own research processes as data is well established in the social sciences and STS (e.g., \citealp{soler2014science}, pp.~12-13), and some CC researchers have begun to adopt similar methods (e.g., \citealp{perezyperez2020towards}). \citet{fiske2020lexical} offers an exemplar of how reflexive description can contribute to a field. They describe the process of becoming interested in the concept of feeling “\textit{moved}” and looking for explication in the literature and primary research. By walking through human elements in the research process and the way that social and institutional factors shaped conclusions, \citeauthor{fiske2020lexical} offers novel commentary on their discipline’s methods while describing their own findings. Ultimately, “polyglotism of [their] research group \ldots helped protect [them] from the lexical fallacy of conflating the usage of a vernacular lexeme–say, \textit{be moved}–with the features of a mental state” (\citeyear{fiske2020lexical}, p.~97). In a similar way, \textbf{improved legitimacy of reflexive data in computer science and human-inspired AI may articulate hidden methodological decision making for researchers in AI and also provoke productive novel discussions. }

\section{Implicit Levels of Analysis}

We have suggested thus far that reflecting on the processes of reading, selecting, and interpreting ideas from other disciplines offers an opportunity to CC researchers working with concepts from human literatures. In this section, we close with an example of an influential idea from human literatures that was present implicitly or explicitly in many of our interviews. Improved awareness of this idea’s origins might help CC researchers, and more broadly AI researchers, articulate methodological beliefs and decisions. 

While describing the concepts they work with, many interviewees invoked the idea of “levels” of explanation, description, reduction, understanding, abstraction, or analysis. There is a long history of considering human and machine minds in terms of levels (see for example, \citealp{putnam1975philosophy}; \citealp{marr1976understanding}; \citealp{poggio2012levels}; \citealp{schouten2007mind}; \citealp{macdougallshackleton2011levels}). In psychology, the use of “levels” is tied to the question of whether higher level constructs in the “mind” can be reduced to physical processes in the brain, beginning with early psychoneural reductionists (Feigl, 1958; Place, 1956; Smart, 1959 in \citealp[p.~6]{schouten2007mind}). This use is therefore deeply tied to the history of cognitive science and AI. Furthermore, such levels are used more broadly in philosophy of science to consider the question of reductionism across disciplines and share a common intellectual history if not identical terminology (\citealp{schouten2007mind}). 

Interestingly, while some interviewees explicitly referenced levels of analysis and cited the computational version described by \citeauthor{marr1976understanding} (\citeyear{marr1982vision}, \citeyear{marr1976understanding}) either in conversation or in scholarly work (e.g., \citealp{wiggins2020creativity}, p.~2; \citealp[pp.~2,~55]{dasgupta2019algorithmic}), many used the language of levels implicitly, and perhaps unconsciously: “psychologists seem to be willing to treat underlying learning as a mystery, which I think is appropriate, and have their models of it. And that’s what I think we’re doing [in AI], even though as researchers we don’t often realize it” (P17). 

\citet{macdougallshackleton2011levels}, writing on diverse uses of levels of analysis in studies of human and animal behaviour, characterizes the essential difference between levels of analysis as between proximate mechanisms vs. ultimate functions. Similarly, many CC interviewees articulated a difference between evaluating creativity based only on an output, like a musical composition, or evaluating creativity based on some aspect of process or mechanism (\citealp{jordanous2016four}, p.~1). 

Meta-commentaries on levels of analysis have suggested that inquiry at any level (mechanism or outcome or any variation thereof) is a valid approach but should be specified clearly (\citealp{schouten2007mind}; \citealp{macdougallshackleton2011levels}). They further identify many false debates between process and product (ibid.). Several interviewees either engaged in a process/product debate or acknowledged its presence in CC: for example, P4 described feeling that something was missing when “[another scholar] came up with a set of criteria for how to evaluate the [creativity of the] output of a system. I couldn’t quite accept that because it didn’t take any notice of the process that the system was using.” Another articulated the importance of distinguishing between \textit{ex post }and \textit{ex ante }definitions in CC, where \textit{ex post }creativity means that creativity is realized when, once achieved, you “can’t tell the difference any more” between the type of mind responsible for an outcome (P12)—expressed by \citet[p.~291]{putnam1975philosophy}, “we could be made of Swiss cheese and it wouldn’t matter.” This conversation is also present in CC literature, with \citet{pease2011computational} distinguishing between “(i) judgements which determine whether an idea or artefact is valuable, and (ii) judgements to determine whether a system is acting creatively” (p.~72, cf. \citealp{wiggins2021computational}; \citealp{hodson2017creative}). 

This example of the levels of analysis lends support to our hypothesis that reflection on the intellectual history of influential ideas from human literatures in CC could lend additional clarity, in this case, to debates about definitions of creativity. Researchers engaging in reflexive commentary on their own work, as suggested in the previous section, might consider documenting the level of analysis they are working with (and whether they intend to remain on that level or take an intentionally integrationist approach; see also \citealp{poggio2012levels}). With respect to our project, we have found that levels of description are implicit in our ability to articulate our scope. This project is concerned with where different researchers’ definitions/understandings of concepts like creativity come from and how they influence subsequent thinking; at what levels of analysis are our project's scope-defining concepts, like creativity, understood? Sharpening our terminology is an ongoing project for us, so we recognize its difficulty, but we also found that levels of analysis continue to help us understand how interviewees formed, implemented, and evaluated their concepts. Indeed, one interviewee explicitly expressed that it would be “really valuable” for AI researchers to “have a little bit more understanding of things like \citeauthor{marr1982vision}’s levels of analysis” (P10). 

\section{Conclusion and Future Work}
 
This paper argues that sharper attention to human elements in research can help generate novel perspectives for CC, and human-inspired AI research more generally. It explores the challenges of interdisciplinary research, suggesting that attention to translation work (including what is being ported across disciplines and how) is beneficial to CC. However, this paper has several limitations: primarily, we present early stage findings. As this project proceeds, we will refine our discussions through additional data collection, seeking theoretical saturation. The term \textit{theoretical saturation} refers to the point at which gathering more data no longer yields further theoretical insights \citep[p.~611]{bryant2007discursive}. For example, we expect that salient new advances like the release of OpenAI’s ChatGPT will contribute to further development of this project, as data collection for this study primarily took place before November 2022. Furthermore, we intend to expand the interview sample to seek psychologists' perspectives on ideas from human literatures that seem to be highly influential in AI and on whether the representations of those ideas in AI are representative of current discussions in psychology. 

Finally, future work will expand the diversity of experience represented by interviewees. In addition to being researchers, interviewees were often artists and had biographical elements informing their work. Rich individual histories influence what researchers see as valuable and what brings them satisfaction or joy, and therefore what is worth studying. Personal narratives and their role in scientific discovery were raised by interviewees (for example, P7 told a story of drawing inspiration from losing a game of ‘memory,’ or ‘concentration,’ to their child) and these narratives clearly play a role in methods and approach. For interviewees in computational creativity, the influence of personal values often included a commitment to uplift rather than replace human creativity (P22, for example, discussed avoiding imitating “\textit{Big-C }creativity” in part because of the potentially severe ethical consequences of doing so). This paper has touched on biography and personal choice, and future work may further develop these themes. 

While this paper has sought to outline some of the human influences on scientific research progress, it does not claim that they can or should be eliminated: rather, that elucidating them will help prompt clearer reflection how human-inspired AI is shaped. As P13 commented, “I’ve learned early on that the best research comes from our lived experience and intuition about the world, and once you have some hypotheses then you apply the scientific method and do things properly, but it’s motivated by our own experiences and that’s where I see the best work getting done. I don’t think we can research something that we don’t live.” 

\section{Acknowledgements}
\footnotesize{We are grateful to numerous contributors, not limited to M.~Ackerman, D.~Brown, W.~Carvalho, J.~Casey, C.~Colas, G.~Cottrell, I.~Dasgupta, J.~Droppo, R.~French, K.~Gordon, C.~Guckelsberger, K.~Gupta, M.~Guzdial,  A.~Jordanous, A.~Lampinen, K.~Mathewson, A.~McLafferty, A.~Parrish, R.~Pérez y Pérez, P.~M.~Pilarski, R.~Saunders, R.~Shariff, W.~Tanenbaum, J.~Virani, G.~Wiggins, and M.~Wilson. NMA was funded by the Canada CIFAR AI Chairs program and the Alberta Machine Intelligence Institute (Amii).}
\vspace{-0.59ex}

{\small \setlength{\bibhang}{0pt}
\bibliographystyle{iccc}
\bibliography{iccc}
}

\end{document}